\documentclass{article}
\usepackage{spconf,amsmath,graphicx}

\usepackage{amsthm, amsfonts, amssymb, mathrsfs, amsmath}
\usepackage{color}
\usepackage{mathtools}
\usepackage{multirow}
\usepackage{subfigure}

\usepackage[ruled,linesnumbered]{algorithm2e}

\newcommand{\bx}{{\boldsymbol{x}}}
\newcommand{\bh}{{\boldsymbol{h}}}
\newcommand{\bu}{{\boldsymbol{u}}}

\newcommand{\bb}{\boldsymbol{b}}
\newcommand{\bs}{\boldsymbol{s}}

\title{Designing Recurrent Neural Networks by unfolding \\ an L1-L1 minimization algorithm}
\name{Hung Duy Le, Huynh Van Luong, Nikos Deligiannis}
\address{Vrije Universiteit Brussel, Pleinlaan 2, B-1050 Brussels, Belgium\\
        imec, Kapeldreef 75, B-3001 Leuven, Belgium
}

\begin{document}

%
\maketitle
\begin{abstract}
We propose a new deep recurrent neural network (RNN) architecture for sequential signal reconstruction. Our network is designed by unfolding
the iterations of the proximal gradient method that solves the $\ell_1\text{-}\ell_1$ minimization
problem. As such, our network leverages by design that signals have a sparse representation and that the difference between consecutive signal representations is also sparse. 
We evaluate the proposed model in the task of reconstructing video frames from compressive measurements and show that it outperforms several state-of-the-art RNN models. 

\end{abstract}
\section{Introduction}
\label{introduction}

The problem of reconstructing sequential signals from low-dimensional---and possibly corrupted---observations across time appears in various imaging applications, including compressive video sensing~\cite{BaraniukSPM17}, dynamic magnetic resonance imaging~\cite{Weizman15}, and mm-Wave imaging~\cite{BecquaertSensors18}. 
When reconstructing time-varying signals, one needs to leverage prior knowledge; namely that (\textit{i}) at a given time instance the signal has a low-complexity representation, such as sparsity in a learned dictionary or fixed basis, and (\textit{ii}) signals (or their representations) across time are correlated (temporal correlation). 

Various methods for sequential signal reconstruction have been proposed in the past. The method in~\cite{Vaswani2008KalmanFC} adapted a Kalman filter in sequential compressed sensing, whereas the Modified-CS method~\cite{zhan2015time} integrates an estimate the signal'�s support into the reconstruction scheme. Alternatively, the methods in \cite{charles2011sparsity,MotaTSP17}
considered that two consecutive
 sparse signal representations are close under an $\ell_1$- or $\ell_2$-norm metric. Such approaches, however, recover the signals using iterative optimization algorithms,  leading to high computational complexity when the dimensionality of the problem increases.
 
%

Deep neural networks (DNNs) have recently achieved state-of-the-art performance in solving inverse problems~\cite{lucas2018using}. These approaches come with the additional benefit of fast reconstruction as they do not have to solve an optimization problem during inference. However, DNNs are black-box models, meaning that they do not integrate prior or domain knowledge, and thus lack interpretability and theoretical guarantees~\cite{lucas2018using}. Recent efforts on designing DNNs that incorporate domain knowledge, include \textit{deep unfolding} methods, which interpret a DNN as an unrolled version of an iterative optimization algorithm. Examples include the learned-ISTA (LISTA) network~\cite{GregorICML10}, which unfolds the iterative soft-thresholding algorithm (ISTA)~\cite{daubechies2004iterative}, the unfolded versions of the approximate message passing~\cite{borgerding2017amp} and iterative hard thresholding~\cite{xin2016maximal} algorithms, and the ADMM-Net~\cite{sun2016deep}.

Little attention has, however, been devoted to the design of \textit{deep recurrent neural networks} (RNNs)~\cite{PascanuICLR14} for representing sequential signals. The authors of~\cite{WisdomICASSP17} proposed an RNN design---named SISTA-RNN---by unfolding the sequential version of ISTA. The problem that the algorithm solves considers that two consecutive signal realizations are close in the $\ell_2$-norm sense. 
 In this work, we propose a novel RNN\ model for sequential signal recovery. Our model is derived by unfolding a proximal gradient method that solves the $\ell_1\text{-}\ell_1$ minimization problem~\cite{charles2011sparsity,MotaTSP17}. This problem assumes that the different between sequential sparse signal representations is also sparse, and it is proven to outperform both $\ell_1$ and $\ell_1\text{-}\ell_2$ minimization~\cite{MotaTIT17}.
We apply the proposed model in the problem of video reconstruction from low-dimensional measurements, that is, sequential frame compressed sensing. Experimentation on the moving MNIST dataset\footnote{The source code to replicate our experiments is available on https://github.com/dhungle.}\cite{SrivastavaICML15} shows that the proposed model achieves higher reconstruction results compared to various state-of-the-art RNN\ models, including SISTA-RNN~\cite{WisdomICASSP17}.

The paper is organized as follows. Section~\ref{background} presents the background of the work. Section~\ref{proposedRNN-l1-l1} describes the proposed model for video reconstruction. Section~\ref{experiment} presents the experiments and section \ref{conclusion} concludes the work.

\section{Background and Related Work}
\label{background}

\subsection{Sparse signal reconstruction} 
\label{sec:sequentialReconstruction}

Consider the problem of reconstructing a sparse signal $\bs\in\mathbb{R}^{n}$ from noisy measurements $\bx=\mathbf{A}\bs+\boldsymbol{\eta}$, where $\mathbf{A}\in \mathbb{R}^{ m\times n}~(m\ll n)$ is a sensing matrix and $\boldsymbol{\eta}$ a noise vector. By leveraging that $\bs$ has a sparse representation $\bh\in\mathbb{R}^{d}$ in a dictionary~$\mathbf{D}\in \mathbb{R}^{ n\times d}$, that is, $\bs = \mathbf{D} \bh$, the signal can be recovered from the measurements by solving~\cite{DonohoTIT06}:
\begin{equation}\label{l1-norm}
\min_{\bh} 
\frac{1}{2}\|\bx -\mathbf{A}\mathbf{D}\bh\|_2^2 + \lambda\|\bh\|_{1},
\end{equation}
where $\|\cdot\|_p$ is the $\ell_{p}$-norm and $\lambda$ is a regularization parameter. ISTA \cite{daubechies2004iterative} solves~\eqref{l1-norm}
by iterating over
\begin{equation}
\label{eq:ista}
\bh^{(i)} = \phi_{\frac{\lambda}{a}}(\bh^{(i-1)} - \frac{1}{a}\mathbf{D}^T\mathbf{A}^T(\mathbf{A}\mathbf{D}\bh^{(i-1)}-\bx)),
\end{equation}
where $\phi_\gamma(u) = \mathrm{sign}(u)(0,|u|-\gamma)_+ $ is the soft thresholding
operator [see Fig.\ref{soft-l1}], $\gamma=\frac{\lambda}{a}$, and $a$ is an upper bound on the Lipschitz
constant of the gradient of $\frac{1}{2}\|\bx -\mathbf{A}\mathbf{D}\bh\|_2^2$. LISTA~\cite{GregorICML10}
unrolls the iterations of ISTA into a feedforward neural
network\ with shared weights, where each
layer implements an iteration: 
$
\bh^{(i)} = \phi_{\gamma}(\mathbf{\Xi}\bh^{(i-1)} + \mathbf{Z}\bx),
$
with $\mathbf{\Xi} = \mathbf{I}-\frac{1}{a}\mathbf{D}^T\mathbf{A}^T\mathbf{A}\mathbf{D}$, $\mathbf{Z} = \frac{1}{a}\mathbf{D}^T\mathbf{A}^T$, and $\gamma$ learned
from data.

\subsection{Stacked-RNNs for sequential signal representation} \label{stack-rnn}

RNNs are connectionist models with self-feedback loops allowing information to pass across sequential steps. In a stacked-RNN~\cite{PascanuICLR14}, see Fig.~\ref{fig:stackedRNN}, the vertical stack of network layers capture the latent representation of an input signal at a given time instance and the horizontal connections learn the temporal relationship across signals.
RNNs can be used to recover a sequence of signals $\bs_t$,
$t=1,2,\dots,$ from a sequence of noisy measurement vectors $\bx_t=\mathbf{A}\bs_t+\boldsymbol{\eta}_t$.
Specifically, given $\bx_{t}$, $t=1,2,\dots$, the signal representation at the $k^{th}$ layer of $K$ layers, $\bh_t^{(k)}$, and the reconstructed signal, $\hat{\bs_t}$ at time~$t$ are calculated as
\begin{equation}\label{stackedRNN}
        \bh_t^{(k)}\hspace{-2pt}=\hspace{-2pt}\left\{
        \begin{array}{l}
         \phi\Big(\mathbf{W}^{(1)}\bh_{t-1}^{(1)}+\mathbf{U}\bx_t\Big),~~~~~~~~~~~~k=1,\\
        \phi\Big(\mathbf{W}^{(k)}\bh_{t-1}^{(k)}+\mathbf{S}^{(k)}\bh_{t}^{(k-1)}\Big),~k>1,\\
        \end{array}
        \right.
        \end{equation}
        \begin{equation}\label{stackedRNN-Output}
        \hat{\bs_t}=\mathbf{V}\bh_t^{(K)} + \bb_t,
\end{equation}
where $\phi(\cdot)$ is a nonlinear activation function such as the \texttt{tanh} or the \texttt{ReLU}, $\mathbf{W},\mathbf{S},~\mathbf{U},~\mathbf{V}$ are affine transformations, and $\bb_t$ are output bias vectors. These parameters, along with initial hidden states $\bh_0$, can be trained using gradient descent with backpropagation-through-time. Different RNN architectures can be applied to solve the problem, e.g., the long short-term memory (LSTM) network~\cite{hochreiter1997long} and the gated recurrent unit (GRU)~\cite{cho2014learning}, which have been proposed to prevent the vanishing gradient problem with long input sequences.

\subsection{The SISTA stacked-RNN\ network}
\label{sec:SISTAstackedRNN}
Traditional RNN models~\cite{hochreiter1997long,cho2014learning,PascanuICLR14} do not integrate the knowledge that the signals $\bs_t$ have sparse representations~$\bh_t$, $t=1,2,\dots$.
 To address this issue, the study in \cite{WisdomICASSP17} proposed a stacked-RNN architecture, which stems from unfolding an iterative soft-thresholding algorithm---referred to as SISTA---that solves the following problem:
\begin{equation}\label{eq:sistaproblem}
        \min_{\bh_{t}} \frac{1}{2}\|\bx_t-\mathbf{A}\mathbf{D}\bh_t\|_2^{2} + \lambda_1\|\bh_t\|_{1} + \frac{\lambda_2}{2}\|\mathbf{D}\bh_t - \mathbf{F}\mathbf{D}\bh_{t-1}\|_{2}^{2},
\end{equation}
where $\mathbf{F}\in \mathbb{R}^{n\times n}$ is a correlation matrix between $\bs_t$ and $\bs_{t-1}$, and $\lambda_1,\lambda_2$ are regularization parameters. Similar to LISTA~\cite{GregorICML10}, the SISTA-RNN network~\cite{WisdomICASSP17} uses the soft-thresholding operator $\phi_\gamma(u)$ [Section~\ref{sec:sequentialReconstruction} and Fig.\ref{soft-l1}] as activation function.

\section{The proposed RNN via $\ell_1\text{-}\ell_1$ minimization}
\label{proposedRNN-l1-l1}
This section describes the proposed RNN\ model, which stems from unfolding the steps of a proximal method that solves the $\ell_1\text{-}\ell_1$ minimization problem. 
\subsection{$\ell_1$-$\ell_1$ minimization in sequential signal recovery}
In sequential signal recovery, one can recover the signal $\bs_t=\mathbf{D}\bh_t$ from measurements~$\bx_t$   by solving the $\ell_1\text{-}\ell_1$ minimization
problem~\cite{charles2011sparsity,MotaTSP17}:
\begin{equation}
\label{l1-l1minimization}
\min_{\bh_t} \frac{1}{2}\|\bx_t-\mathbf{A}\mathbf{D}\bh_t\|_2^2 + \lambda_1\|\bh_t\|_{1} + \lambda_2\|\bh_t - \textbf{G}\bh_{t-1}\|_{1} ,
\end{equation}
where $\textbf{G}\in \mathbb{R}^{d\times d}$ is an affine transformation that promotes the correlation between the sparse representations between two consecutive instantiations of the signal, $\bh_{t-1}$ and $\bh_{t}$, and $\lambda_1,\lambda_2>0$~are regularization parameters. We highlight that in~\eqref{l1-l1minimization} the correlation between $\bh_{t}$ and $\textbf{G}\bh_{t-1}$ is encoded using the~$\ell_1$-norm as opposed to using the~$\ell_2$-norm to express the correlation between $\bs_{t}$ and $\textbf{F}\bs_{t-1}$ in SISTA [see~\eqref{eq:sistaproblem}]. The motivation is twofold: Firstly, it is proven that~$\ell_1$-$\ell_1$ outperforms $\ell_1$-$\ell_2$ minimization in sparse signal recovery~\cite{MotaTIT17}. Secondly, from an application perspective, we know that the error between consecutive video frames (or their motion-compensated versions) typically follow the Laplace rather than the Gaussian distribution~\cite{deligiannis2014maximum}.   
\begin{algorithm}[t]
        \textbf{Input:} Sequence of measurements~$\bx_{1:T}$, measurement matrix~$\textbf{A}$, dictionary~$\textbf{D}$, affine transformation~$\textbf{G}$, initial sparse code~${\bh}_0^{(K)}$, parameters $\alpha$, $\lambda_1$, $\lambda_2$.\\
        \For{t = 1,\dots,T}{
         $\bh_t^{(1)} = \textbf{G}{\bh}^{(K)}_{t-1}$\\
         \For{k = 1 to K}{
            $\bu = [\textbf{I} - \frac{1}{\alpha} \textbf{D}^T\textbf{A}^T
\textbf{AD}]\bh_t^{(k-1)} + \frac{1}{\alpha}\textbf{D}^T \textbf{A}^T \bx_t$\\
                $\bh_t^{(k)} = \phi_{\frac{\lambda_1}{\alpha}, \frac{\lambda_2}{\alpha}, \textbf{Gh}^{(K)}_{t-1}}(\bu)$\\
        }
        }
        \textbf{return} $\bh_{1:T}^{(K)}$\\
        \caption{The proposed proximal gradient method for sequential signal recovery via $\ell_{1}$-$\ell_{1}$ minimization.}
        \label{sista-algorithm}
\end{algorithm}

\begin{figure}[t]
        \centering
        \subfigure[]{\label{soft-l1}\includegraphics[width=0.15\textwidth]{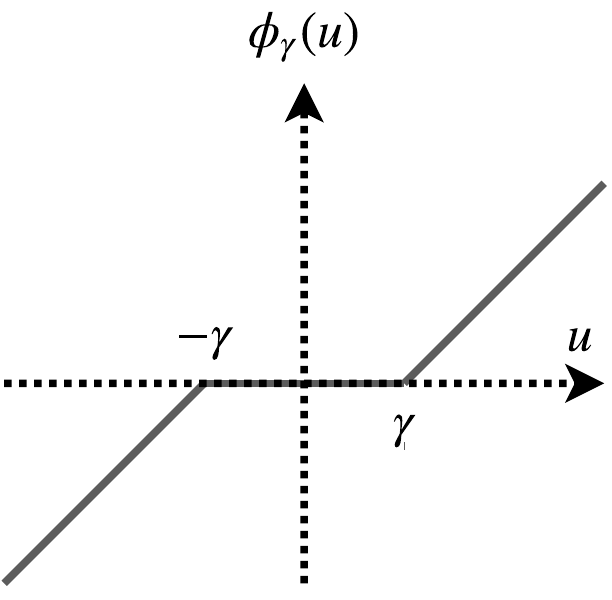}}
        \subfigure[]{\label{soft-l1-l1}\includegraphics[width=0.24\textwidth]{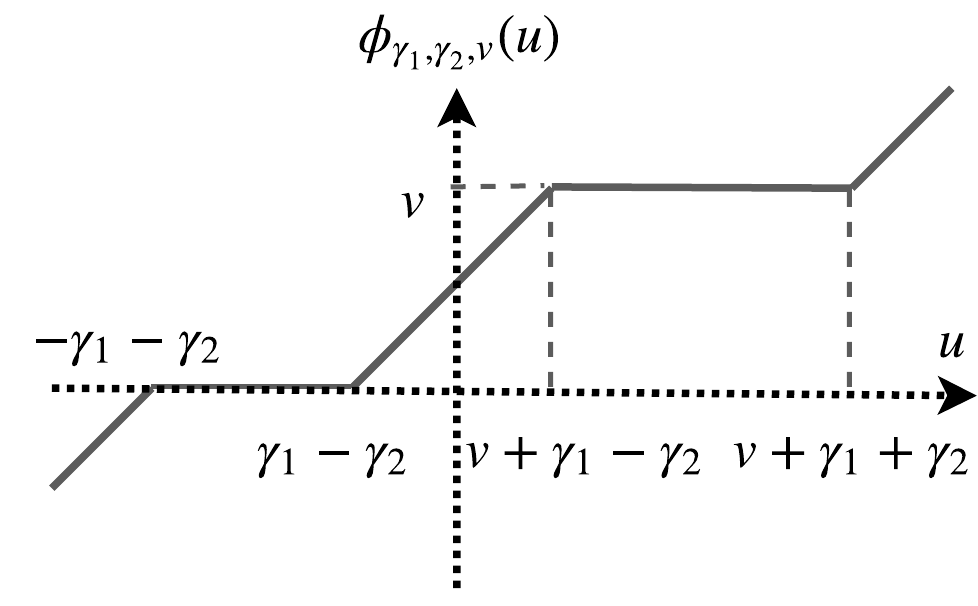}}
        \caption{The $\ell_{1}$ and $\ell_{1}$-$\ell_1$ soft-thresholding
functions ($v \geq 0$).}
        \label{soft-function}
\end{figure}

The objective function of~\eqref{l1-l1minimization} consists of two parts: the differentiable function $f(\bh_t) = \frac{1}{2}\|\bx_t-\mathbf{A}\mathbf{D}\bh_t\|_2^{2}$ and the non-smooth function $g(\bh_t) = \lambda_1\|\bh_t\|_{1}+\lambda_2\|\bh_t - \textbf{G}\bh_{t-1}\|_{1}$. Hence, in this work, we propose to solve~\eqref{l1-l1minimization} using a proximal gradient method, the steps of which are given in Algorithm~\ref{sista-algorithm}. In our algorithm, step 5 applies a gradient descent update for~$f(\bh_t)$ with a learning rate~$\frac{1}{a}>0$ and step 6 applies element-wise the proximal operator for our problem, $\phi_{\frac{\lambda_1}{\alpha},\frac{\lambda_2}{\alpha},\textbf{Gh}^{(K)}_{t-1}}(\bu)$. For notation brevity, we denote~$\gamma_1=\frac{\lambda_1}{\alpha}$, $\gamma_2=\frac{\lambda_2}{\alpha}$ and~$u$, $v$ an element in~$\bu$,~$\textbf{Gh}^{(K)}_{t-1}$, respectively; the proximal operator is then defined as:
\begin{align}
\phi_{\gamma_1, \gamma_2, v}(u) \hspace{-2pt}=\hspace{-2pt}
\begin{cases}
u\hspace{-1pt} -\hspace{-1pt} \gamma_1\hspace{-1pt} - \hspace{-1pt}\gamma_2, &\hspace{-1pt} v \hspace{-1pt}+ \hspace{-1pt}\gamma_1 \hspace{-1pt}+ \hspace{-1pt}\gamma_2 \hspace{-1pt}\leq u \hspace{-1pt}< \hspace{-1pt}\infty \\
v,                       &\hspace{-1pt} v \hspace{-1pt}+ \hspace{-1pt}\gamma_1 \hspace{-1pt}- \hspace{-1pt}\gamma_2 \hspace{-1pt}\leq u \hspace{-1pt}<\hspace{-1pt} v\hspace{-1pt} + \hspace{-1pt}\gamma_1\hspace{-1pt} +\hspace{-1pt} \gamma_2 \hspace{-1pt}\\
u\hspace{-1pt} - \hspace{-1pt}\gamma_1 \hspace{-1pt}+ \hspace{-1pt}\gamma_2, &\hspace{-1pt} \gamma_1 - \gamma_2 \leq u < v + \gamma_1
- \gamma_2 \\
0,                       &\hspace{-1pt} -\gamma_1 - \gamma_2\leq u < \gamma_1- \gamma_2\\
u \hspace{-1pt}+ \hspace{-1pt}\gamma_1 \hspace{-1pt}+ \hspace{-1pt}\gamma_2, &\hspace{-1pt} -\infty < u < -\gamma_1 - \gamma_2 \\
\end{cases}
\end{align}
if $v\geq 0$, and 
\begin{align}\phi_{\gamma_1, \gamma_2, v}(u) \hspace{-2pt}=\hspace{-2pt}
\begin{cases}
u\hspace{-1pt} - \hspace{-1pt}\gamma_1\hspace{-1pt} - \hspace{-1pt}\gamma_2, & \hspace{-1pt}\gamma_1 \hspace{-1pt}+ \hspace{-1pt}\gamma_2 \hspace{-1pt}\leq \hspace{-1pt}u \hspace{-1pt}<\hspace{-1pt} \infty\\
0,                        & \hspace{-1pt}-\hspace{-1pt}\gamma_1\hspace{-1pt} + \hspace{-1pt}\gamma_2\hspace{-1pt} \leq u \hspace{-1pt}<\hspace{-1pt} \gamma_1 \hspace{-1pt}+ \hspace{-1pt}\gamma_2\\
u \hspace{-1pt}+ \hspace{-1pt}\gamma_1 \hspace{-1pt}- \hspace{-1pt}\gamma_2,   &\hspace{-1pt} v \hspace{-1pt}- \gamma_1 \hspace{-1pt}+ \hspace{-1pt}\gamma_2 \hspace{-1pt}\leq u\hspace{-1pt} <\hspace{-1pt} -\hspace{-1pt}\gamma_1 \hspace{-1pt}+ \hspace{-1pt}\gamma_2\\
v,                         & \hspace{-1pt}v\hspace{-1pt} - \hspace{-1pt}\gamma_1\hspace{-1pt} \hspace{-1pt}- \hspace{-1pt}\gamma_2 \hspace{-1pt}\leq u \hspace{-1pt}< \hspace{-1pt}v\hspace{-1pt} -\hspace{-1pt} \gamma_1\hspace{-1pt} + \hspace{-1pt}\gamma_2 \\
u \hspace{-1pt}- \hspace{-1pt}\gamma_1 \hspace{-1pt}+ \hspace{-1pt}\gamma_2,   &\hspace{-1pt} -\hspace{-1pt}\infty \hspace{-1pt}< \hspace{-1pt}u\hspace{-1pt} <\hspace{-1pt} v\hspace{-1pt} -\hspace{-1pt} \gamma_1\hspace{-1pt} -\hspace{-1pt} \gamma_2,\\
\end{cases}
\end{align}
if $v < 0$. 
Fig.~\ref{soft-function} depicts a schema of the proximal operator of our algorithm for $v\geq 0$ [see Fig.~\ref{soft-l1-l1}] in comparison with the soft-thresholding operator~$\phi_\gamma(u)$
[see Fig.~\ref{soft-l1}], which is used in SISTA~\cite{WisdomICASSP17}.

\begin{figure}[t]
        \centering
        \subfigure[A stacked RNN.]{\label{fig:stackedRNN}\includegraphics[width=0.27\textwidth]{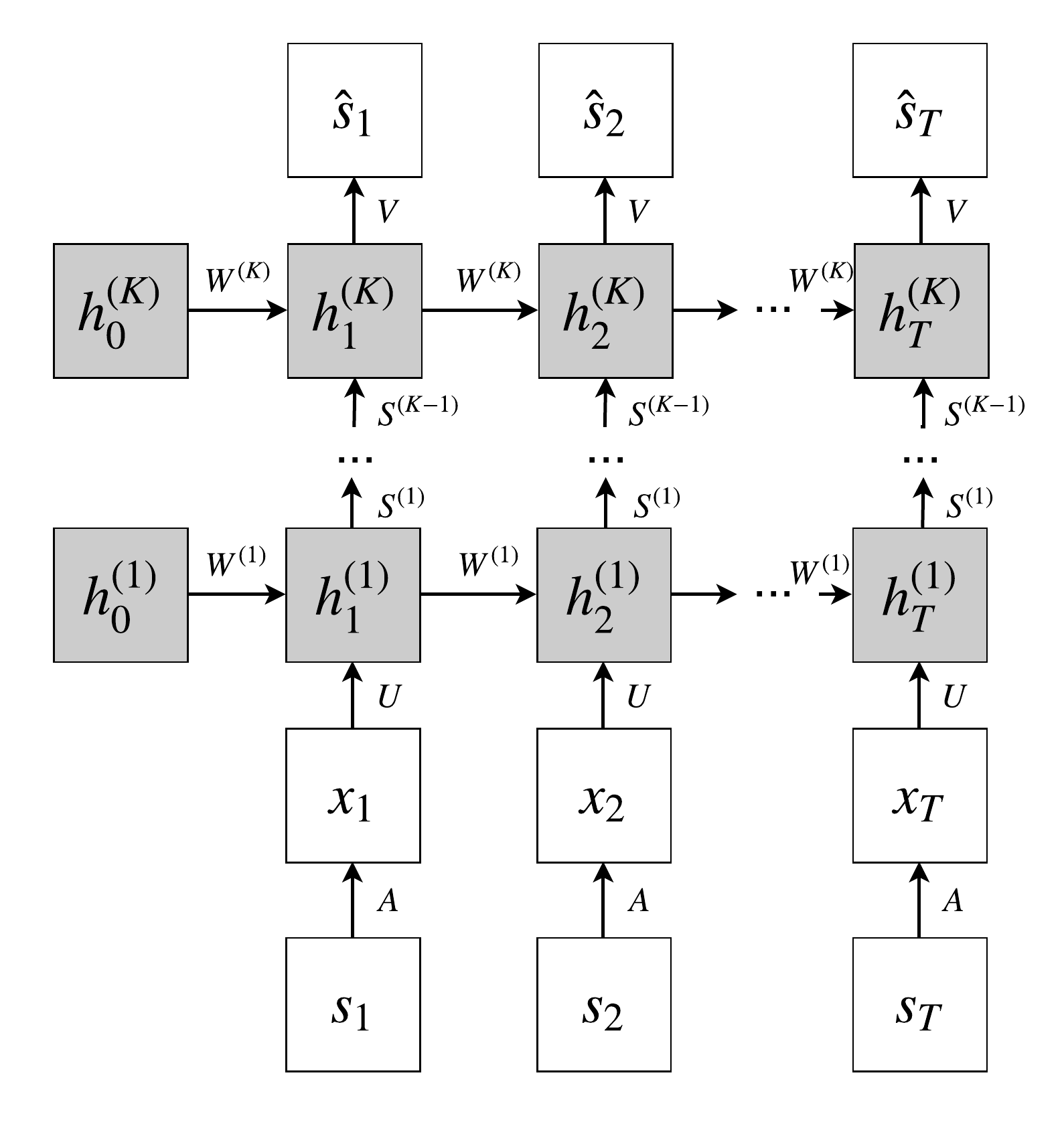}}

        \subfigure[The proposed model.]{\label{proposed-model}\includegraphics[width=0.28\textwidth]{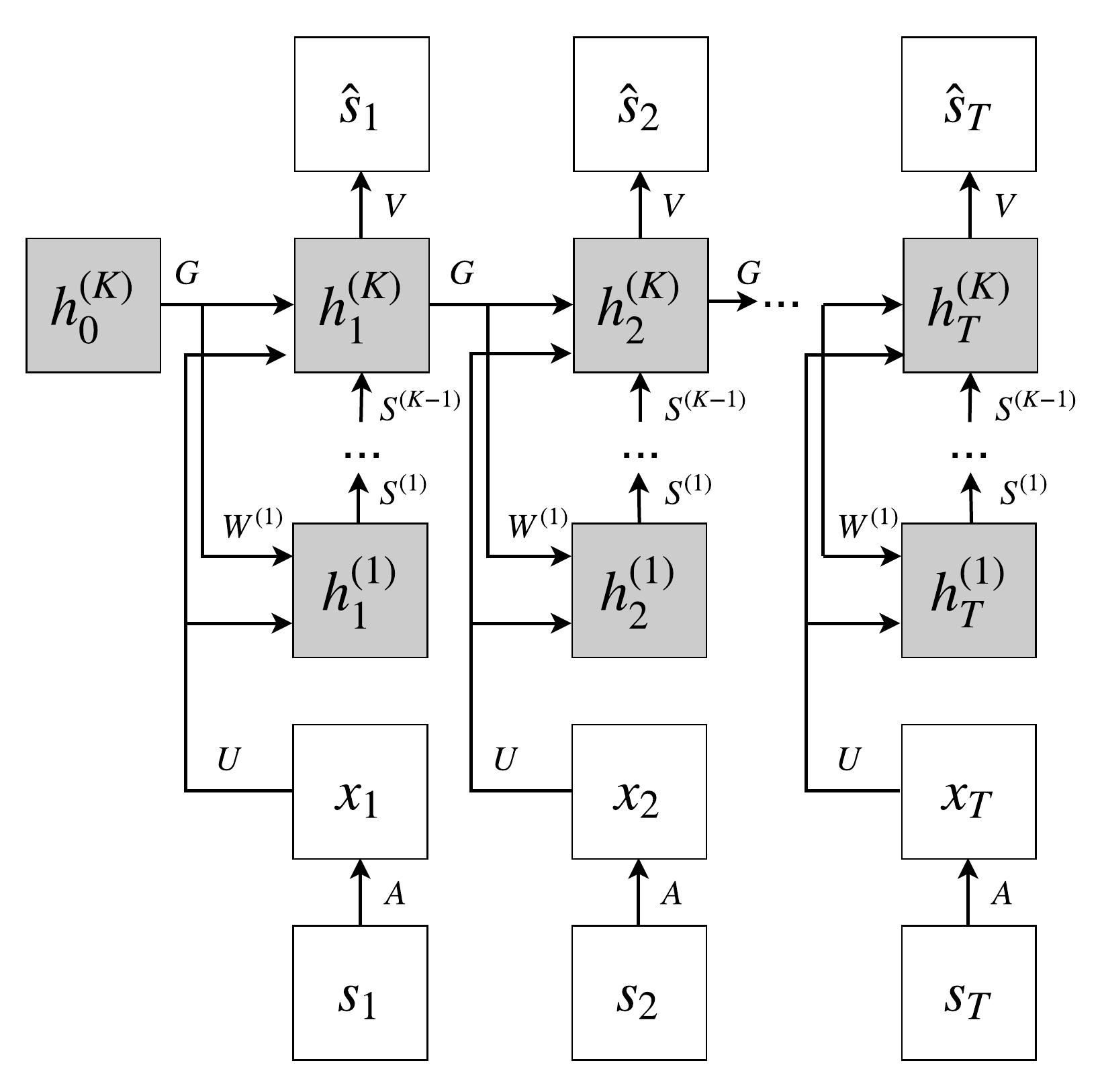}}

        \caption{The proposed RNN vs. a stacked RNN.}
        \label{network}
\end{figure}

\subsection{The proposed $\ell_{1}$-$\ell_1$-RNN architecture}
We now describe the proposed stacked-RNN\ architecture for sequential signal recovery, which we call~$\ell_{1}$-$\ell_1$-RNN. The network, which is shown in Fig.~\ref{proposed-model}, is designed by unrolling the steps of Algorithm \ref{sista-algorithm} across the iterations $k = 1,\dots,K$ (yielding the hidden layers) and time instances $t = 1,\dots,T$.
Specifically, the $k$-th hidden layer is given by 
\begin{equation}\label{stackedRNN}
\bh_t^{(k)}\hspace{-2pt}=\hspace{-2pt}\left\{
\begin{array}{l}
\phi_{\frac{\lambda_1}{\alpha}, \frac{\lambda_2}{\alpha}, \textbf{Gh}^{(K)}_{t-1}}\Big(\mathbf{W}^{(1)}\bh_{t-1}^{(1)}+\mathbf{U}\bx_t\Big),~\text{if}~k=1,\\
\phi_{\frac{\lambda_1}{\alpha}, \frac{\lambda_2}{\alpha}, \textbf{Gh}^{(K)}_{t-1}}\Big(\mathbf{S}^{(k)}\bh_{t}^{(k-1)} + \textbf{U}\bx_t\Big),~\text{if}~k>1,\\
\end{array}
\right.
\end{equation}
and the reconstructed signal at time instance~$t$ is calculated as
\begin{equation}\label{stackedRNN-Output}
\hat{\bs_t}=\mathbf{V}\bh_t^{(K)} + \bb_t.
\end{equation}
where $\mathbf{U}$, $\mathbf{W}$, $\mathbf{S}^{(k)}$, $\mathbf{V}$ are defined as
\begin{align}
        &\mathbf{U}=\frac{1}{\alpha}\mathbf{D}^T\mathbf{A}^T, \forall k, \\
        &\mathbf{W}^{(1)}=\textbf{G} - \frac{1}{\alpha} \textbf{D}^T\textbf{A}^T \textbf{AD}\textbf{G},\\
        &\mathbf{S}^{(k)}=\mathbf{I}-\frac{1}{\alpha}\mathbf{D}^T\mathbf{A}^T\mathbf{A}\mathbf{D}, ~k>1,\\
        &\mathbf{V}=\mathbf{D},~\bb_t=\mathbf{0}.
\end{align}
The activation function has the form of the proximal operator $\phi_{\gamma_1, \gamma_2, v}(u)$ with the parameters $\gamma_1$, $\gamma_2$ learned during training.
We train our network in an end-to-end fashion: Vectorized frames are inputs $\bs_t$, $t=1,\dots,T$, which are compressed by a linear measurement layer $\textbf{A}$, resulting in compressive measurements $\bx_{t}$. The reconstructed frames $\hat{\bs}_t$ are obtained by multiplying linearly the hidden representation $\bh_{t}^{(K)}$ with the dictionary $\textbf{D}$. During training, we minimize the loss function~$\mathcal{L} = \sum_{t=1}^{T}\|\bs_t - \hat{\bs_t}\|_{2}^{2} + \beta\|\mathbf{\theta}\|_{2}^{2}$ using Adam optimization \cite{kingma2014adam} on mini-batches, where the trainable parameters are $\mathbf{\theta}= \{\textbf{A}, \textbf{D}, \textbf{G}, \bh_0, \alpha, \lambda_1, \lambda_2\}$ 
and $\beta$ is the hyper-parameter of the weight decay regularization. 

\section{Experiments}
\label{experiment}
We assess the performance of the proposed RNN model in the problem of video frame reconstruction from compressive measurements. We use the moving MNIST dataset~\cite{SrivastavaICML15} for our experiments; the 10.000 video sequences (with 20 frames each) in the dataset are split into non-overlapping training, validation, and test sets, consisting of 8.000, 1.000, and 1.000 sequences, respectively. In order to reduce the computational complexity and memory requirements, each frame has been spatially downscaled from $64\times
64$ to $16\times 16$ pixels using bilinear decimation. After vectorising, we obtain signal sequences of $\bs_{1:T}\in \mathbb{R}^{256}$ with $T=20$ time steps. For each sequence, we obtain a sequence of measurements $\bx_{1:20}$ using a trainable linear sensing matrix $\textbf{A}\in \mathbb{R}^{m\times n}$, with~$n = 256$ and $m < n$. We test several different values of $m$ corresponding to compression rates of $\{50\%, 33\%, 25\%, 20\%\}$. The dictionary $\textbf{D}\in \mathbb{R}^{256\times 1024}$ is initialized with the overcomplete discrete cosine transform (DCT). 

\begin{table}[t]        
\label{results}
        \caption{Average PSNR results [in dB] for sequential frame reconstruction
on the testing dataset.}
        \label{tab:result}
        \centering
        \begin{tabular}{| c || c | c | c | c |}
                \hline
                \multirow{2}{*}{Model} & \multicolumn{4}{ c |}{Compression rate}\\ 
                \cline{2-5}
                 &50\% & 33\% & 25\% & 20\%\\
                 \hline \hline
                Stacked-RNN &  38.56 & 35.82 & 33.12 & 30.78 \\  
                Stacked-LSTM  &  37.02 & 34.06 & 31.55 & 29.60 \\  
                Stacked-GRU &  40.3 & 37.31 & 33.98 & \textbf{31.09} \\ 

                SISTA-RNN &  42.52 & 37.20 & 33.85 & 30.91\\
                \hline
                Our model &  \textbf{44.65} & \textbf{38.90} & \textbf{34.22}
& 30.76\\
                \hline
        \end{tabular}
\end{table}

We compare the reconstruction performance of the proposed RNN model against existing various RNN\ models, namely, SISTA-RNN~\cite{WisdomICASSP17}, stacked-RNN~\cite{pascanu2013construct}, stacked-LSTM (with the LSTM cell architecture from~\cite{hochreiter1997long}), and stacked-GRU (with the GRU cell architecture from~\cite{cho2014learning}). Following the experimental setup in~\cite{WisdomICASSP17}, we initialize the sparse code $\bh_0^{(K)}$ as the zero vector and $\alpha = 1.0$. We empirically found that using a weight decay $\beta=0.01$ in the SISTA-RNN and the proposed $\ell_1$-$\ell_1$-RNN yields best results in the validation set, whereas for the rest of the models not using a weight decay ($\beta=0$) provides best results. In order to initialize $\lambda_1,\lambda_2$ [see \eqref{l1-l1minimization}], we perform a random search to find the best combination that yields high reconstruction accuracy in the validation dataset, and obtain $\lambda_1 = 1.0$, $\lambda_2 = 0.01$ for our model and $\lambda_1 = 0.3$, $\lambda_2 = 0.01$ for SISTA-RNN. We use $K =3$ stacks for all models, except for the stacked-LSTM model\footnote{The stacked-LSTM\ with $K=3$ could not be trained from the data, possibly due to the large amount of parameters to optimise.} for which we set $K = 2$. All weights and biases are initialized with a uniform distribution of $\pm\frac{1}{\sqrt{d}}$, where $d=1024$ is the size of each hidden layer. We train the networks for 200 epochs with a learning rate of 0.0003, and a batch size of 32.

\begin{figure}[t]
        \centering
        \includegraphics[width=0.49\textwidth]{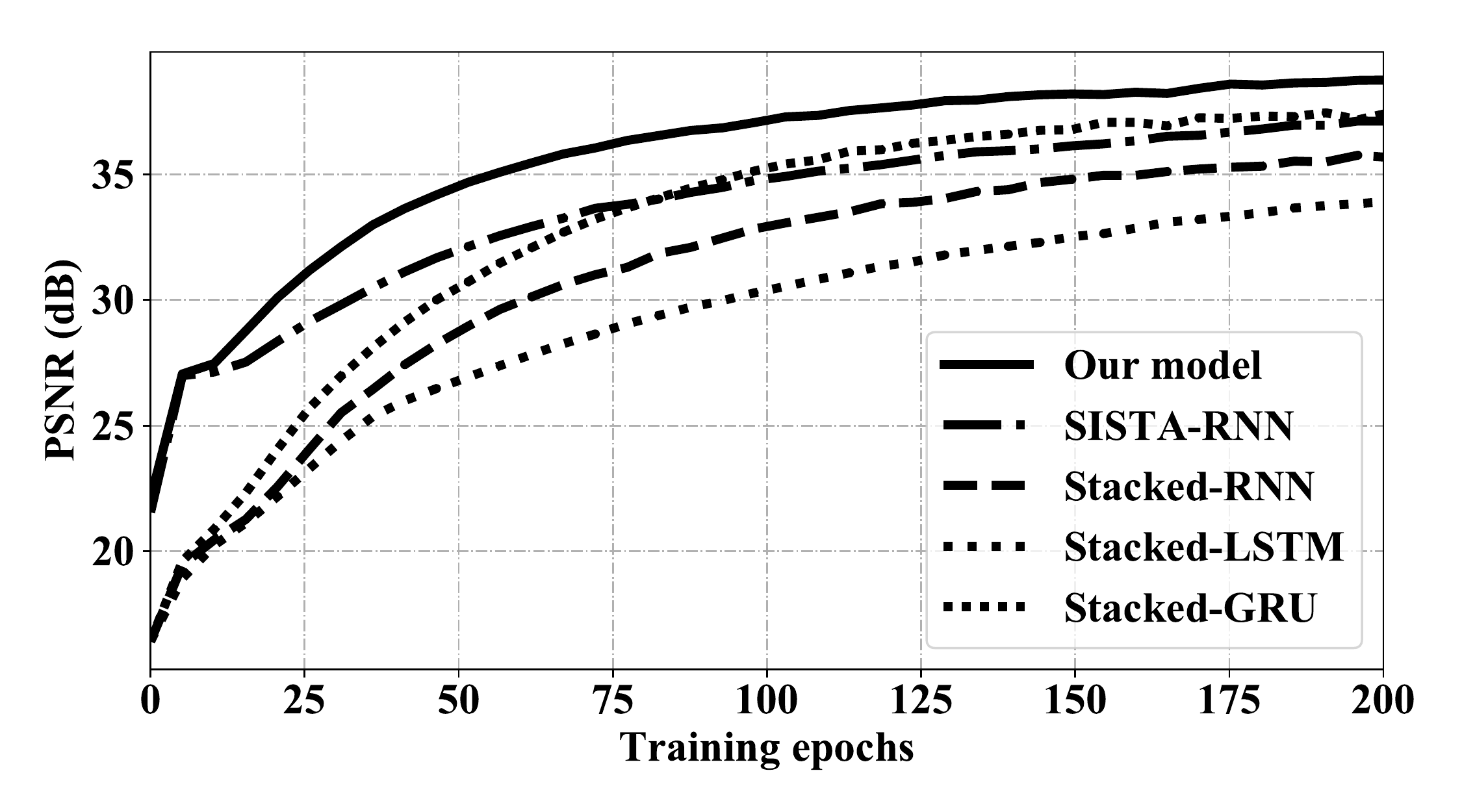}
        \caption{Learning curves for all RNN\ models measured using the PSNR calculated on the validation
set under a compression rate of $33\%$.}
        \label{fig:losscurve}
\end{figure}

Table~\ref{tab:result} reports the reconstructed PSNR results averaged across all frames and sequences in the test set. The experiments show that our model outperforms all other models at compression rates $50\%, 33\%$, and $25\%$, bringing respective improvements of 2.13~dB, 1.59~dB, 0.24~dB over the second best model. At the rate of $20\%$, the proposed model is outperformed by SISTA-RNN and stacked-GRU. 
In addition, Fig.~\ref{fig:losscurve} illustrates average PSNR (measured
on the validation set) versus training epochs curves for all models at a compression rate of~$33\%$. The learning curve of the proposed RNN is consistently better than those of the other models. 

After 200 epochs of training our model, we measured the sparsity (number of zero elements) of 55\% in the last layer $\bh_t^{(3)}$. We leave the investigation of how sparsity affects reconstruction performance for future work.

\section{Conclusion}
\label{conclusion}
We proposed a stacked RNN for sequential sparse signal recovery from compressive measurements. Our RNN architecture incorporates prior knowledge about the structure of the signals and their correlation by deep-unfolding a proximal gradient method for the $\ell_1\text{-}\ell_1$ minimization problem. Our experiments in the task of video-frame recovery from compressive measurement show that our model outperforms several state-of-the-art RNNs. 

\bibliographystyle{IEEEbib}
\bibliography{refs}

\end{document}